\documentclass[runningheads]{llncs}

\usepackage{graphicx}
\usepackage{caption}
\usepackage{subcaption}
\usepackage{amsfonts}
\usepackage{amssymb, amsmath}

\title{Navigating Conceptual Space;} 
\subtitle{A new take on Artificial General Intelligence.}
\titlerunning{Navigating Conceptual Space}
\author{Per Roald Leikanger}
\institute{UiT - The Arctic University of Norway,\\
    Tromsø, Norway\\
    \email{Per.R.Leikanger@uit.no}}
\begin{document}   
\maketitle


\begin{abstract}


    Edward C. Tolman found reinforcement learning unsatisfactory for explaining intelligence and proposed a clear distinction between learning and behavior.
    Tolman's ideas on latent learning and cognitive maps eventually led to what is now known as conceptual space, 
        a geometric representation where concepts and ideas can form points or shapes.
    Active navigation between ideas -- reasoning -- can be expressed directly as purposive navigation in conceptual space.
    Assimilating the theory of conceptual space from modern neuroscience, we propose autonomous navigation as a valid approach for emulated cognition.
    However, achieving autonomous navigation in high-dimensional Euclidean spaces is not trivial in technology.
    In this work, we explore whether \mbox{neoRL} navigation is up for the task;
        adopting Kaelbling's concerns for efficient robot navigation, we test whether the \mbox{neoRL} approach is general across navigational modalities, 
        compositional across considerations of experience, and effective when learning in multiple Euclidean dimensions.
    We find neoRL learning to be more resemblant of biological learning than of RL in AI,
        and propose neoRL navigation of conceptual space as a plausible new path toward emulated cognition. 

\end{abstract}

%

\section{Introduction}

Edward C. Tolman first proposed cognitive maps for explaining the mechanism behind rats taking shortcuts and what he referred to as latent learning \cite{tolman1948cognitive}.
    Tolman was not satisfied with behaviorists' view that goals and purposes could be reduced to a hard-wired desire for reward \cite{chaplin1961systems}.
Experiments showed that unrewarded rats would perform better than the fully rewarded group when later \emph{motivated} by reward \cite{tolman1930degrees}.
Arguing that a reinforcement signal was more important for behavior than for learning, 
    Tolman proposed the existence of a cognitive model of the environment in the form of a \emph{cognitive map}. 
The mechanisms behind neural representation of Euclidean space (NRES) has later been identified for a range of navigational modalities by electrophysical measurements \cite{bicanski2020neuronal}.
Further, the NRES mechanism has been implied for navigating \emph{conceptual space} \cite{constantinescu2016organizing},
    a Euclidean representation where betweenness and relative location makes sense for explaining concepts \cite{gardenfors2000conceptual}. 
Results from theoretical neuroscience indicate NRES' role in 
social navigation \cite{schafer2018navigating}, temporal representation \cite{eichenbaum2014time}, and \mbox{reasoning \cite{bellmund2018navigating}}.
    Cognitive maps for representing thought have received much attention in neurophysiology in the recent five years 
    \cite{bellmund2018navigating,constantinescu2016organizing,schafer2018navigating}.
    Navigating conceptual space as an analogy of thought could explain generalization and reasoning based on locality \cite{gardenfors2000conceptual}.
%


     \begin{figure}[tbh!p]
         \centering
         \includegraphics[width=0.75\columnwidth]{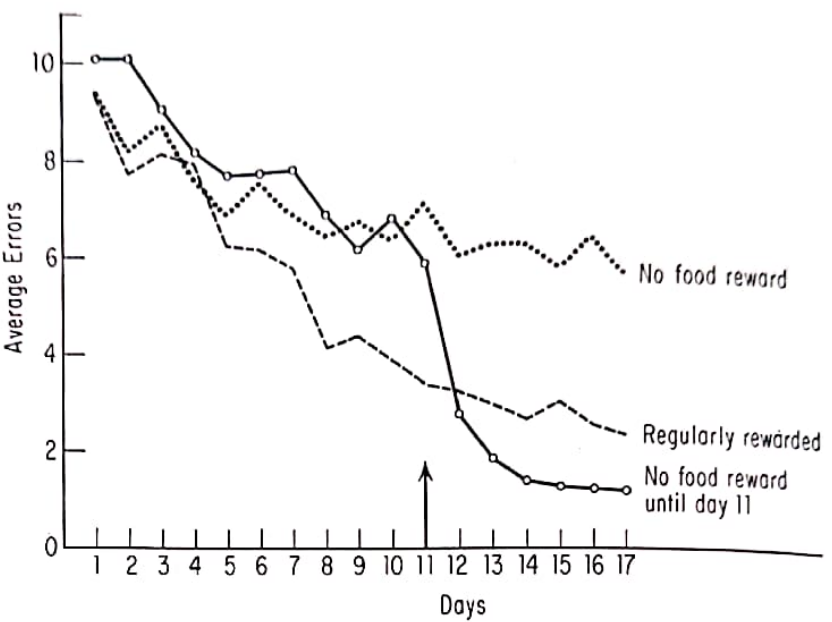}
         \caption{ Evidence for latent learning by {Tolman and Honzik} (1930). \\(after \cite{tolman1930degrees}, from \emph{\small Systems and theories of psychology} \cite{chaplin1961systems}).}
         \label{fig:latent_learning}
     \end{figure}

    Autonomous navigation is difficult to reproduce in technology.
    Autonomous operation implies a decision \emph{agent} capable of forming decisions based on own desires and experience.
    A well-renowned approach to establish experience-based behavior is reinforcement learning (RL) from AI.
    Via trial and error based on a scalar reward signal $\mathbb{R}$, a decision \emph{agent} is capable of adapting behavior according to the accumulation of $\mathbb{R}$.
    Considering robot path planning as Euclidean navigation, we look toward robot learning for inspiration on autonomous navigation.
    However;
    whereas RL powered by deep function approximation has been demonstrated for playing board games at an expert level, 
        requirements to sample efficiency combined with high Markov dimensionality in temporal systems makes deep RL difficult in navigation learning \cite{kober2013reinforcement}.
    Leslie Kaelbling (2020) points out key challenges for efficient robot learning, apparently concerned with the current direction of deep RL.
    Navigation has to be efficient (require few interactions for learning new behaviors), 
        general (applicable to situations outside one's direct experience), 
        and compositional/incremental (compositional with earlier knowledge, incremental with earlier considerations). 
    The current state-of-the-art deep RL for robotics struggles on all three points \cite{kaelbling2020foundation}.

    Inspired by neural navigation capabilities, Leikanger (2019) has developed an NRES-oriented RL (neoRL) architecture for online navigation \cite{leikanger_CCN_2019}.
    Via orthogonal value functions (OVF) formed by off-policy learning toward each cell of an NRES representation, 
        the neoRL architecture allows for a distinction between learning and behavior.
    Inspired by animal psychology, the neoRL framework allows purposive behavior to form based on the desire for anticipated reward \cite{leikanger_ALIFE_2021}.
    However, navigating a multi-dimensional conceptual space of unknown dimensionality in real-time would be impossible for any current learning algorithm. 
    In this work, we adopt Kaelbling's three concerns when testing whether \mbox{neoRL} navigation allows for autonomous navigation in high-dimensional Euclidean space. 
\section{Theory} 
    Central to all navigation is knowledge of one's current navigational state. 
    Information about relative location, orientation, and heading to objects that can block or otherwise affect the path is crucial for efficient navigation.
    When such knowledge is represented as vectors relative to one's current configuration, neuroscientists refer to this representation as being \emph{egocentric}.
    When represented relative to some external reference frame, coordinates are referred to as being \emph{allocentric}.
    Vectors can be expressed as Cartesian coordinates, e.g. the vector $\vec{a} = [1.0, 3.0]$ represent a point or displacement in a plane, 
        one unit size from the origin along the first dimension, and three units along a second dimension.
    Vectors can also be represented in polar coordinates $\vec{a} = [r, \varphi]$, a point with distance $r$ from the origin in the allocentric direction $\varphi$.
    In order to apply RL for navigation, all this information must be represented according to the Markov property;
        each instance of agent state must contain enough data to define next-state distribution \cite{sutton2018reinforcement}.
    Combined with temporal dynamics, the number of such instances becomes prohibitively expensive for autonomous navigation by RL \cite{kober2013reinforcement}.
    Neural state representation, on the other hand, appears to be fully distributed across individual neurons and parts of the hippocampal formation \cite{solstad2009neural}.
    NRES coding for separate navigational modalities (as should be represented in separate Euclidean spaces) have been located in different structures in the hippocampal formation \cite{bicanski2020neuronal}.
    Navigational state representation for the only system capable of true autonomous navigation seems to be decomposed across multiple NRES modalities. 
    This section introduces theory and considerations on how state is represented in the animal and the learning machine, 
        an important inspiration for neoRL mechanisms for navigation and problem solving.

    \begin{figure}[tbh!p]
        \centering

        \includegraphics[width=\columnwidth]{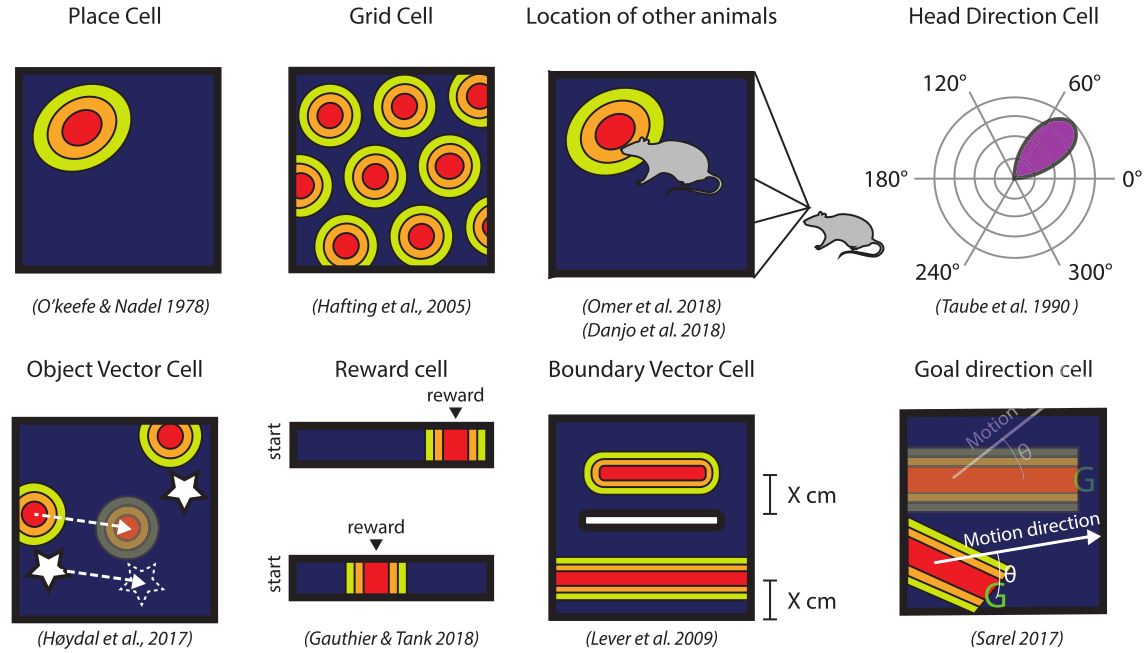}
         \caption{
                    Some identified NRES modalities of importance for navigation, with reference to the original publication.
                    All NRES modalities could be important for autonomous spatial navigation.
                    The place cell and the object vector cell will be of particular importance in the examples and experiments of this text.
                    (Illustration adopted from \cite{behrens2018cognitive} )
        }
        \label{fig:illustration_of_some_NRES}
    \end{figure}

\subsection{Neural Representation of Euclidean Space}
\label{sec:NRES}
    The first identified NRES neuron was the \emph{place cell} \cite{o1971hippocampus}; 
        O'Keefe and Dostrovsky discovered that specific neurons in the hippocampus became active whenever the animal traversed a specific location in the test environment.
    Reflecting the allocentric position of the animal, the individual place cell could be thought of as a geometric \emph{feature detector} on the animal's location;
        the place cell is active whenever the animal is located within the \emph{receptive field} of the cell. 
    Other NRES cells have later been identified, expressing information in various parameter spaces.
    Identified NRES modalities for navigation includes:
            one's allocentric location \cite{o1971hippocampus}, 
            allocentric polar vector coordinates to external objects \cite{hoydal2019object}, 
            and one's current heading \cite{taube1990head}. 
    A selection of relevant NRES modalities is listed in \mbox{Table \ref{table:some_allocentric_NRES_an_overview}} or in \mbox{Figure \ref{fig:illustration_of_some_NRES}}.
    A more comprehensive study on NRES modalities in neurophysiology has been composed by Bugress and Bicanski \cite{bicanski2020neuronal}.

    \begin{table*}[t!hb]
        \small
        \centering
        \begin{tabular}{||l | c | c | c | l | l ||} 
         \hline
                                   &           Location        &       Tuning                     &      Direction        &   NRES modality                &                   \\
         \hline                                                                                                                                                                           
            Place Cell             &           ac.             &       [proximal]  2D             &       -               &   Current position             & \cite{o1971hippocampus}           \\
            Border Cell            &           ac.             &       [proximal]  2D             &       -               &   Location of borders          & \cite{solstad2008:border_cells}   \\
            Object Vector Cell     &           polar c.        &       [spectrum]  2D             &      ac.              &   Location of objects          & \cite{hoydal2019object}           \\
            Boundary Vector Cell   &           polar c.        &       [spectrum]  2D             &      ac.              &   Location of boundaries       & \cite{lever2009boundary}          \\
            Head-Direction Cell    &           -               &       [angular]\quad  1D         &      ac.              &   Head direction               & \cite{taube1990head}              \\
            Speed Cell             &           -               &       [rate code] 1D             &       -               &   Current velocity             & \cite{kropff2015speed}            \\
         \hline
        \end{tabular}
            \vspace{9pt}
        \caption{
            Neural representation for different Euclidean spaces of importance for navigation:
            Head-direction cells reflect the current allocentric (\emph{ac.}) angle of the head (a scalar parameter).
            The place cell and border cell respond to a proximal allocentric location (2D). 
            The remaining NRES reflect conditions represented in other Euclidean spaces -- listed as NRES modalities. 
        }
        \label{table:some_allocentric_NRES_an_overview}
    \end{table*}

\begin{figure}[htb!p]
        \begin{subfigure}[b]{0.3\textwidth}
            \centering
            \fbox{\includegraphics[width=\textwidth]{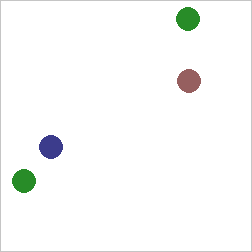}}
        \end{subfigure}
        \hfill
        \begin{subfigure}[b]{0.3\textwidth}
            \centering
            \fbox{\includegraphics[width=\textwidth]{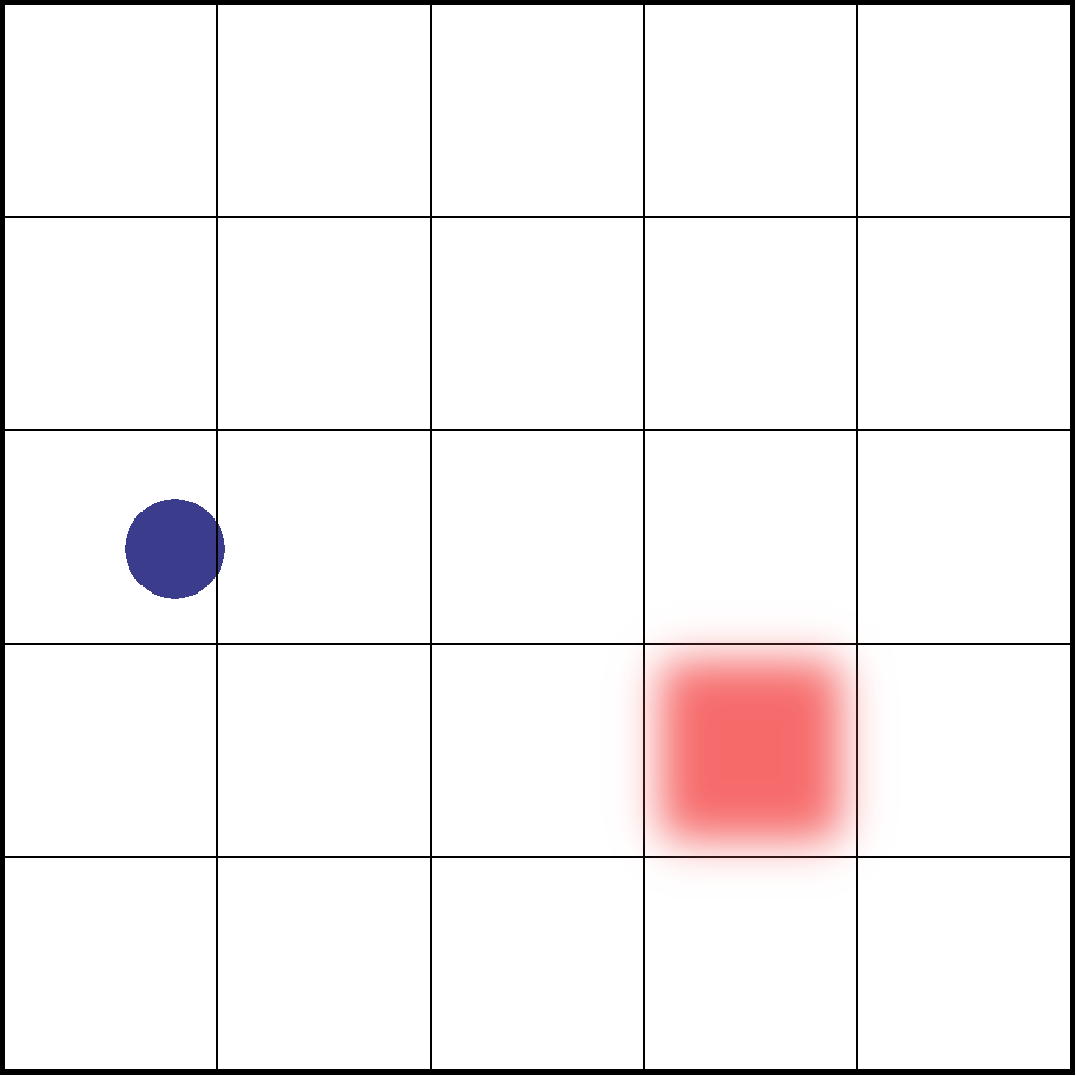}}
        \end{subfigure}
        \hfill
        \begin{subfigure}[b]{0.3\textwidth}
            \centering
            \fbox{\includegraphics[width=\textwidth]{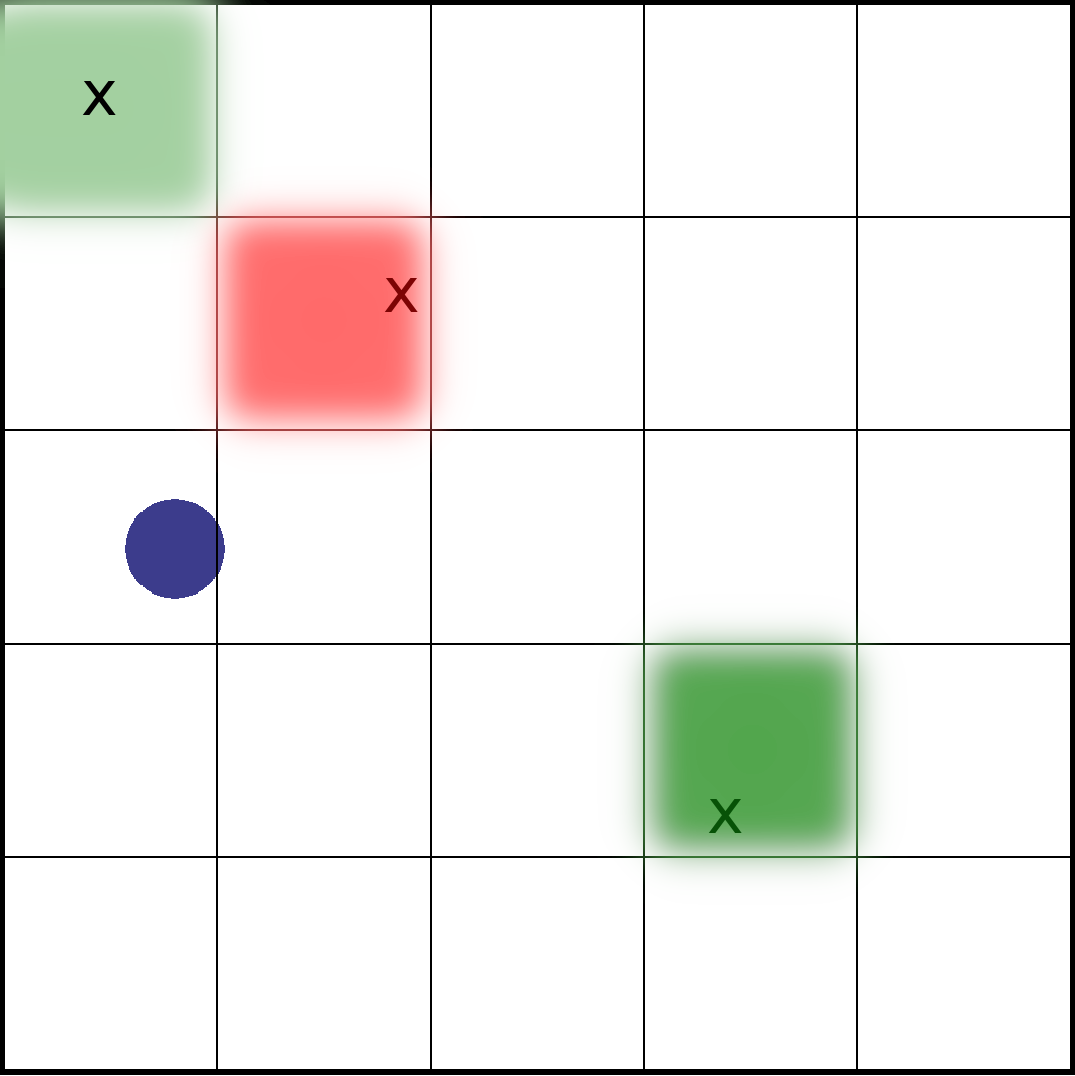}}
        \end{subfigure}
        \caption{
            (A) The allocentric WaterWorld environment: Blue entity is governed by inertia dynamics, with a desire for green ($\mathbb{R}=+1$) and aversion for red ($\mathbb{R}=-1$).
            (B) An $N5$ mapping of NRES: Each axis is divided into $N = 5$ equal intervals, resulting in $N^2 = 25$ NRES cells.
            An OVF represents the value function toward one NRES activation.
            (C) Learned NRES maps can form behaviors via anticipated reward: 
            When an NRES tile contains an element associated with reward, the corresponding OVF is weighted accordingly.
            Anticipated rewards are illustrated using the same colors as in {\tiny(A)};
                one aversive NRES cell in red and two desirable NRES cells associated with various anticipation are represented in shades of green.
        }
        \label{fig:WaterWorld_illustration}
    \end{figure}

    \label{sec:one-hot}
    Neuroscientists assume that populations of NRES neurons map Euclidean vectors by neural patterns of activation.
    A simple mapping could be formed by a population of NRES cells responding to mutually exclusive receptive fields.
    One could visualize this representation as a chessboard;
        exactly one cell (tile) would be satisfied for any point on the board.
    Referred to as \emph{one-hot encoding}\footnote{ \footnotesize
                Note for computing scientists: 
                NRES is not concerned with the Markov state. 
                Any similarity to RL coarse coding and CMAC can therefore be considered to be an endorsement of these AI techniques, not grounds for direct comparison.
        } in computing sciences, a mutually exclusive map structure is defined by the resolution and the geometric coverage of the map's tiles.
    This intuitive map is appropriate for demonstrative purposes:
    All examples and experiments in this text are encoded by a comprehensive one-hot mapping as illustrated in Figure \ref{fig:WaterWorld_illustration},
        where, e.g., $N13$ signifies a $13x13$ tile set in $\Re^2$.
\subsection{Autonomous Navigation by neoRL Agents} 
\label{sec:theoryRL}
    One can separate navigation into two distinct aspects;
        the desired location -- the objective of the interaction -- and how this objective can be reached.
    When both aspects are governed by one's own inclinations and \emph{experience}, we refer to this as an autonomous operation. 
    %
    A most accomplished approach to experience-based behavior is RL from AI;
        a decision \emph{agent} can be thought of as an algorithmic entity that learns how better to reach an objective by trial and error.
    The decision process of the agent can be summarized by 3 signals: 
        the \emph{state} of the system before the interaction, 
        the \emph{action} with which the agent interacts with the system, 
        and a \emph{reward} signal that reflects the success of the operation with regard to an objective. 
    Experience can be expressed via the \emph{value function}, reflecting the expected total reward from this state and forward under the current policy. 
    Since behavior (policy) is based on the current value function, 
        and the value function is defined under one policy, 
        an alternating iterative improvement is required for learning.
    The resulting asymptotic progress is slow, requiring many interactions by RL learning.  
    Although RL has proven effective for solving a range of algorithmic tasks, autonomous control for robotics remains a challenge \cite{kober2013reinforcement}.
    Even RL powered by deep function approximation (deep RL) has limited applicability for online interaction learning in Euclidean spaces \cite{kaelbling2020foundation}.


    A neoRL agent, on the other hand, is composed by a set of sub-agents learning how to achieve different NRES cells for the corresponding NRES representation \cite{leikanger_CCN_2019}.
    The whole set of learning processes constitutes the (latent) learning aspect of the agent;
        behavior can later be harvested as a weighted sum over the OVFs according to priorities \cite{leikanger_ALIFE_2021}.
    Learning OVFs as general value functions \cite{sutton2011horde} with $\mathbb{R}$ defined by NRES cell activation, 
        the value function of the whole neoRL agent resembles Kurt Lewin's \emph{fieldt theory} of learning \cite{lewin1942field}.
    Leikanger (2021) demonstrated how emulated NRES for agent state allows for autonomous navigation in a single Euclidean space \cite{leikanger_ALIFE_2021},
        however, multi-modal navigation and combining experience across NRES modalities remains to be tested.
    %
    As multi-modal NRES capabilities would bring neoRL state representation closer to navigational state representation in the brain, 
        compositionality across NRES modalities would be important for making neoRL a plausible candidate for conceptual navigation.

\section{Multi-modal neoRL navigation}
\label{sec:method}

    Adopting Kaelbling's three concerns for Euclidean navigation, we next explore how neoRL navigation scales with increasing (Euclidean) dimensionality.
    First, it is crucial that NRES-oriented navigation can operate based on different Euclidean spaces; 
        with little knowledge of the form or meaning of conceptual spaces, neoRL must be capable of navigation by other information than location. 
    Further, we are interested in how neoRL navigation scales with additional parameters or across multiple NRES modalities. 
    Any exponential increase in training time with additional states would make conceptual navigation infeasible. 
    \mbox{NeoRL} navigation must be \emph{general} across NRES modalities,
        \emph{compositional} across conceptual components, 
        without any significant decline in learning \emph{efficiency}.
    In this section, we explore neoRL capabilities for hi-dimensional navigation by experiments inspired by Kaelbling's concerns for efficient robot navigation.
    %
    %
    %
       
    All experiments are conducted in the allocentric version of the WaterWorld environment \cite{webWaterWorld_in_PLE}, 
        illustrated in Figure \ref{fig:WaterWorld_illustration}A.
    An agent controls the movement of the self (blue dot), with a set of actions that accelerate the object in the four directions $N$, $S$, $E$, $W$.
    Three objects of interest move freely in a closed section of a Euclidean plane.
    When the agent encounters an object, it is replaced by a new object with a random color, location, and speed vector.
    Green objects are desirable with an accompanying reward $\mathbb{R} = +1.0$, and red objects should be avoided with $\mathbb{R} = -1.0$. 
    No other rewards exist in these experiments, making $\mathbb{R}$ a decent measure of an agent's navigation capabilities. 
    Note that the agent must catch the last green in a board full of red before the board can be reset and continue beyond (on average) 1.5 points per reset.



    The PLE \cite{tasfi2016PLE} version of WaterWorld reports the Cartesian coordinates of the agent and elements of interest;
        thinking of the Euclidean plane in Figure \ref{fig:WaterWorld_illustration}A as representing location facilitates later discussion. 
    A direct NRES encoding of this information will be referred to as place cell (PC) NRES modality in the remainder of this text.
    One can also compute a simple object vector cell (OVC) interpretation by vector subtraction:
        $$ \vec{o^i}_{\text{\tiny OVC}} = \vec{o^i}_{\text{\tiny{PC}}} - \vec{s}_{\text{\tiny{PC}}} $$
    where $\vec{s}$ is the location of self and $\vec{o^i}$ is the location of object $i$ in $PC$ or $OVC$ reference frame.
    Note that this OVC interpretation allows for a modality similar to OVC with the self in the center and allocentric direction to external objects, but not with polar coordinates as reported for OVC \cite{hoydal2019object}.
    However, the two Cartesian representations of location still give different points of view due to different reference frames.
    %
    %
    Information is encoded in NRES maps as described in section \ref{sec:one-hot};
        the neoRL agent is organized across multiple NRES maps of different resolutions as described in \cite{leikanger_ALIFE_2021}.
    Multi-res NRES modalities cover resolutions given by primes up to $N13$, i.e., with layers $N2, N3, N5, N7, N11, \text{and }  N13$.
    For more on multi-resolution neoRL agents and the mechanism behind policy from parallel NRES state spaces, see \cite{leikanger_ALIFE_2021}.
    All execution runs smoothly on a single CPU core, and the agent starts with no priors other than described in this section.
    Referring to the NRES modalities as PC and OVC for WaterWorld is only syntactical to facilitate later discussions; 2D Euclidean coordinates are general and can represent any parameter pair. 
    Learning efficiency is compared by considering the transient proficiency of the agent as measured by the reward received by the agent during $0.2 s$ intervals.
    Any end-of-episode reward is disabled in the WaterWorld settings; the only received reward is $\mathbb{R}=+1$ when encountering green elements and $\mathbb{R}=-1$ when encountering red elements.
    The simple reward structure makes accumulated $\mathbb{R}$ a direct measure of how well the agent performed during one run.
    However, observing the transient proficiency -- real-time learning efficiency -- of the agent requires further analysis:
        in all experiments, a per-interval average or received reward is computed over 100 independent runs with additional smoothing by a Butterworth low-pass filter.
    All runs are conducted in isolation;
        the agent is initiated before each run and deleted after the run -- without any accumulation of experience between runs.
    The x-axis of every plot represents minutes since agent initiation. 
    The y-axis represents proficiency as computed by the per-interval average of received reward, scaled to reflect [$\mathbb{R}/s$].
    Proficiency thus measures how many more desirable (green) encounters happen per second than unwanted (red) ones.

\subsection{NeoRL navigation: NRES generality} 
    First, we examine the generality of the neoRL architecture by comparing navigational proficiency for an agent exposed to a PC modality to one exposed to an OVC modality.
    %
    We are interested in the generality of neoRL navigation; can \mbox{neoRL} navigate the PC modality by different Euclidean information, and at what cost?
    %

    \begin{figure}[htb!p]
        \begin{subfigure}[b]{0.5\textwidth}
            \hfill
            \includegraphics[width=\textwidth]{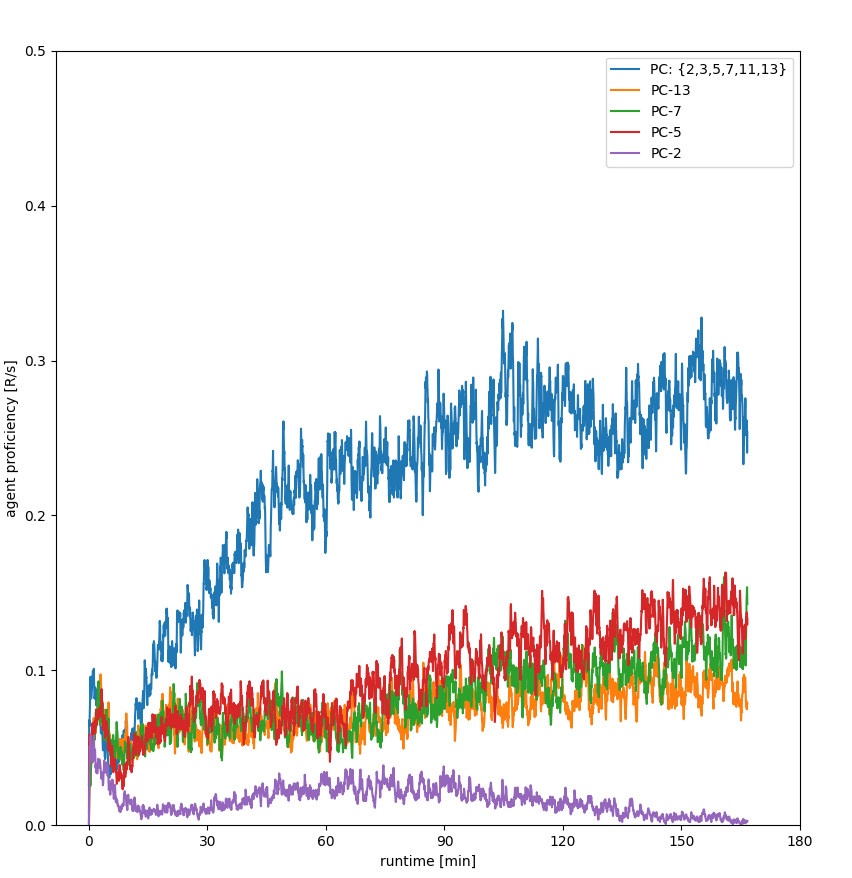}
        \end{subfigure}
        \begin{subfigure}[b]{0.5\textwidth}
            \hfill
            \includegraphics[width=\textwidth]{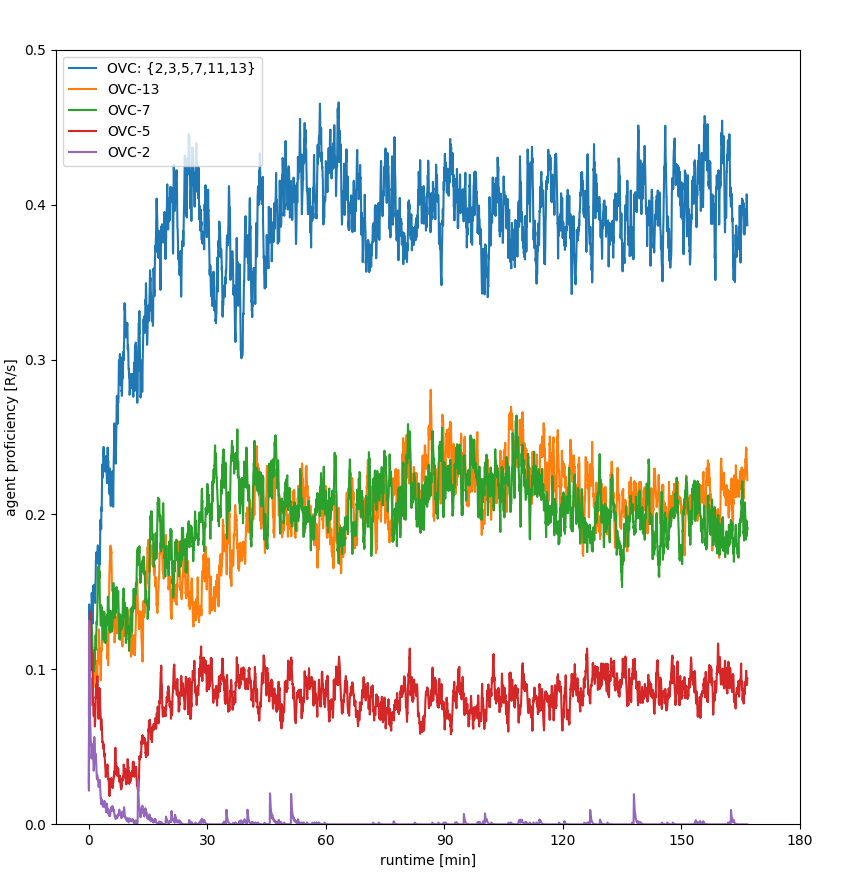}
        \end{subfigure}

        \caption{ 
            The neoRL architecture is general across NRES modalities:
            (A) an original place cell (PC) NRES modality, implemented by applying NRES code directly on an allocentric location of the agent or elements of interest.
            (B) an emulated object vector cell (OVC) NRES modality, implemented by vector subtraction.
            OVC is centered on the self with an allocentric representation of other objects. 
        }
        \label{fig:results:generality}
    \end{figure}

    Results are presented in Figure \ref{fig:results:generality}: 
        agent proficiency from the original PC modality (Figure \ref{fig:results:generality}A) can be compared with agent proficiency when navigating by the OVC modality (Figure \ref{fig:results:generality}B).
    The immediate proficiency of several mono-resolution neoRL agents is plotted alongside the proficiency of a multi-resolution neoRL agent.
    There is no loss in sample efficiency when utilizing the OVC modality compared to PC modality.
    The multi-res neoRL agent performs better than mono-res neoRL agents for both the PC and the OVC modality. 
    %
    NeoRL navigation performs well across both aspects of experience, indicating that the neoRL architecture is general across navigational modalities.

\subsection{NeoRL navigation: NRES compositionality} 
    Secondly, we are interested in how neoRL scales with additional navigational information.
    %
    Experiment 1 showed how a neoRL agent is capable of reactive navigation based on an auxiliary NRES modality.
    In this experiment, we explore the benefit of combining experience across more than one NRES modality.
    A multi-modal neoRL agent is exposed to both the PC and the OVC modality from experiment 1, effectively doubling the number of NRES states for the agent to consider.
    We are anxious about how well the neoRL architecture scales with the additional information, both for final proficiency and learning time. 


    \begin{figure}[htb!p]
        \includegraphics[width=1.0\textwidth]{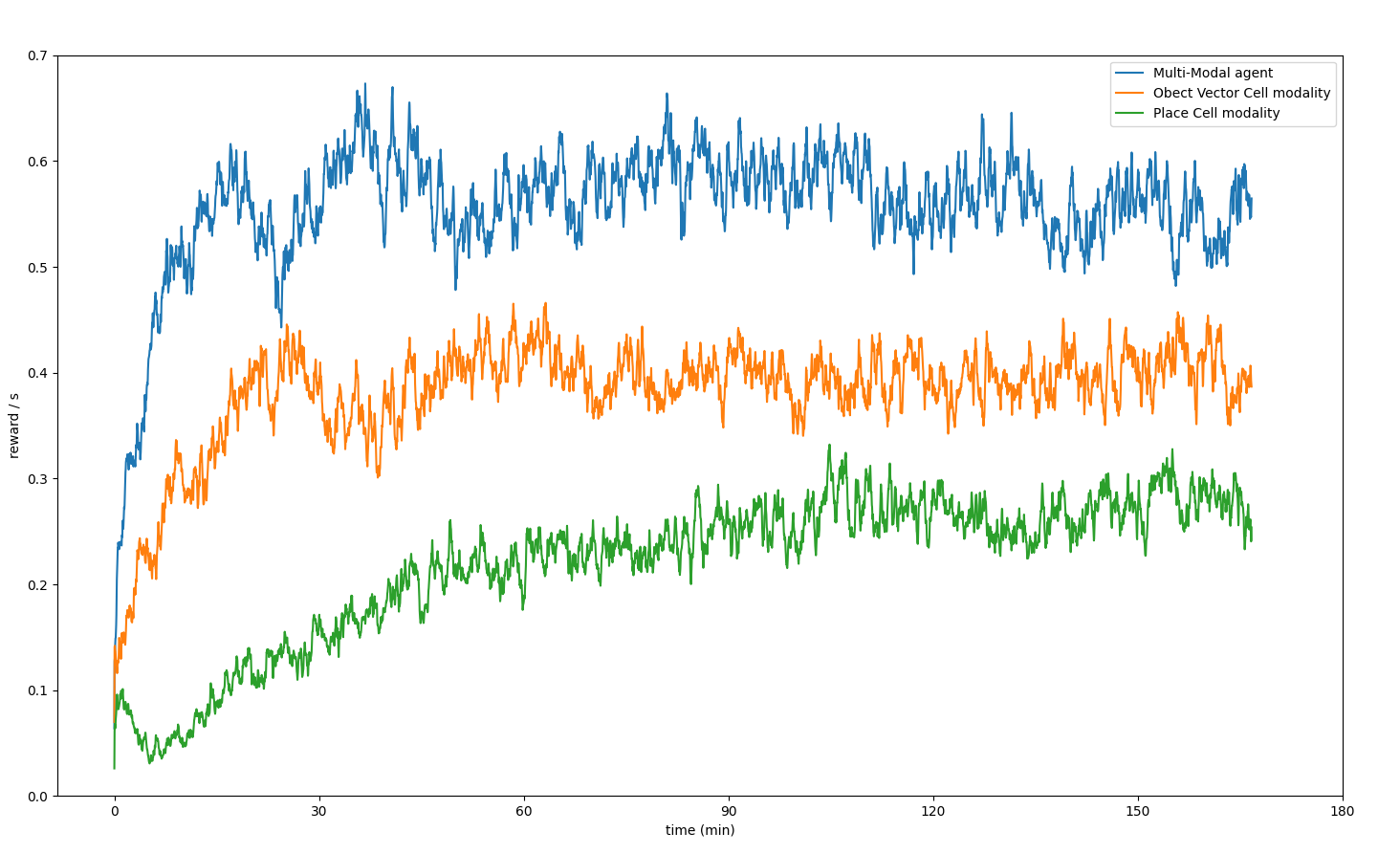}
        \caption{ 
            Multi-modal neoRL navigation leads to higher proficiency and quicker learning than mono-modal agents, despite having twice as many NRES states. 
        }
        \label{fig:results:compositioality}
    \end{figure}

    Compare the proficiency of the neoRL agent when exposed to PC, OVC, and multi-modal information in Figure \ref{fig:results:compositioality}.
    Combining information across multiple NRES modalities significantly improves navigational performance.
    %
    The final proficiency of the multi-modal neoRL agent approaches $0.55 [\mathbb{R}/s]$ while the PC neoRL agent barely reaches $0.29 [\mathbb{R}/s]$.
    The multi-modal neoRL agent reaches final proficiency after $15$ minutes, whereas the PC neoRL agent uses more than $160$ minutes.
    In terms of learning efficiency, i.e., how fast the agent reaches final proficiency, and in terms of trained performance, 
        the multi-modal neoRL agent performs better than both mono-modal neoRL agents.
\section{Discussion}
\label{sec:discussion}

    Contrary to RL in AI, 
        neoRL navigation learns quicker, to higher proficiency, when more information is available to an agent.
    The neoRL agent is capable of multi-modal navigation, making multi-dimensional Euclidean navigation by a digital agent plausible.
    Moving on from reinforcement learning and classical behaviorism, 
        Tolman made a clear distinction between learning and performance after his latent learning experiments (see Figure \ref{fig:latent_learning}).
    Observing how an animal could learn facts about the world that could subsequently be used in a flexible manner, Tolman proposed what he called purposive behaviorism.
    When motivated by the promise of reward, the animal could utilize latent knowledge to form beneficial behavior toward that objective.
    %
    %
    %
    %
    Mechanisms underlying orientation have further been implied in cognition,
        a conceptual space where ideas are represented as points in a multi-dimensional Euclidean space.
    Technological advances have allowed new evidence from modern neuroscience, supporting Tolman's hypotheses on cognitive maps' involvement in thought.
    Inferring that active navigation of such a space corresponds to reasoning and problem solving, 
        we here propose autonomous navigation of conceptual space as an interesting new approach to artificial general intelligence. 
    However, navigating conceptual space -- with high dimensionality, an unknown form, and possibly an evolving number of Euclidean dimensions, is no trivial challenge for technology.
    Based on neural representation of space, the neoRL architecture is distributed and concurrent in learning, capable of separating between latent learning and purposive behavior, 
        and a good candidate for emulated cognition by autonomous navigation of conceptual space.


    Adopting Kaelbling's concerns for efficient robot learning to account for multi-modal navigation, we have methodically tested neoRL navigation in the WaterWorld environment.
    Firstly, it is crucial that neoRL navigation can operate in other Euclidean spaces than its primary navigation modality.
    Our first experiment verifies that the neoRL architecture is general across Euclidean spaces;
        a neoRL agent that navigates by the location modality is compared to one exposed to a relative-vector representation of external objects.
    Both NRES modalities perform admirably at this task, indicating that neoRL navigation is not restricted to one NRES modality.
    Secondly, we explore how neoRL navigation scales with additional NRES modalities; 
        an agent based on both a place-cell and an object-vector-cell representation is compared to the two mono-modal neoRL agents from experiment 1.
    Navigation, both in training efficiency and final proficiency, improves significantly when more information is available to the agent. 
    High-dimensional Euclidean navigation appears to be plausible with neoRL technology, formed by the basic principles from neuroscience and NRES.
    In this work, we have collected evidence from theoretical neuroscience and the psychology of learning to propose a new direction toward emulated cognition. 
    We have shown how online autonomous navigation is feasible by the neoRL architecture;
        still, the most interesting steps toward conceptual navigation in machines remain.
    What are the implications of autonomous navigation of conceptual space for AGI?
    Could latent spaces from deep networks be used for neoRL navigation?
    Should desires (elements of interest) propagate across NRES modalities based on associativity?
    Many important questions are yet to be asked.
    In showing that neoRL is up for the task of multi-modal navigation, we hereby propose a novel approach to AGI and present a plausible first step toward conceptual navigation in machines.
\bibliographystyle{splncs04}
\bibliography{peerlearning}

\end{document}